\documentclass[11pt]{article}
\usepackage{acl2015}
\usepackage{times}
\usepackage{url}
\usepackage{latexsym}
\usepackage{color}
\usepackage{float}
\usepackage{amsmath,amsfonts,amssymb}
\usepackage{algorithm}
\usepackage{appendix}
\usepackage{algpseudocode}
\usepackage{graphicx,subfigure}
\usepackage{booktabs}
\usepackage{multirow}
\usepackage{tikz,pgfplots}
\usepackage{enumitem}

\usepackage[font=small]{caption}

\makeatletter
\def\BState{\State\hskip-\ALG@thistlm}
\makeatother

\pgfplotscreateplotcyclelist{mark list*}{%
every mark/.append style={solid,fill=.!80!black},mark=*\\%
every mark/.append style={solid,fill=.!80!black},mark=square*\\%
every mark/.append style={solid,fill=.!80!black},mark=triangle*\\%
every mark/.append style={solid,fill=.!80!black},mark=halfsquare*\\%
every mark/.append style={solid,fill=.!80!black},mark=pentagon*\\%
every mark/.append style={solid,fill=.!80!black},mark=halfcircle*\\%
every mark/.append style={solid,fill=.!80!black,rotate=180},mark=halfdiamond*\\%
every mark/.append style={solid,fill=.!80!black!40},mark=otimes*\\%
every mark/.append style={solid,fill=.!80!black},mark=diamond*\\%
every mark/.append style={solid,fill=.!80!black},mark=halfsquare right*\\%
every mark/.append style={solid,fill=.!80!black},mark=halfsquare left*\\%
}
\pgfplotscreateplotcyclelist{color list}{%
red,blue,black,yellow,brown,teal,orange,violet,cyan,green!70!black,magenta,gray}
\pgfplotscreateplotcyclelist{linestyles}{solid,dashed,dotted}
\pgfplotscreateplotcyclelist{linestyles*}{solid,dashed,dotted,dashdotted,dashdotdotted}

\tikzset{every mark/.append style={scale=1.5}}

\pgfplotscreateplotcyclelist{mylist}{%
{blue,mark=*,solid},
{green!60!black,mark=triangle},
{brown!60!black,mark=diamond*},
{brown!60!black,mark options={fill=brown!40},mark=otimes*},
{yellow!60!black,mark=triangle*,mark options={fill=brown!40}},
{red,mark=square,}}

\newcommand{\mb}[1]{\mathbf{#1}}
\newcommand{\tb}[1]{\textbf{#1}}

\DeclareMathOperator*{\argmin}{arg\,min}
\DeclareMathOperator*{\argmax}{arg\,max}

\newcommand{\eg}{{\it e.g.}}

\newcommand{\ie}{{\it i.e.}}

\newcommand{\sysname}{\textsc{S-mart}}

\enlargethispage{\baselineskip} 

\title{\sysname: Novel Tree-based Structured Learning Algorithms\\ Applied to Tweet Entity Linking}

\author{Yi Yang\\
  School of Interactive Computing \\
  Georgia Institute of Technology \\
  {\tt yiyang@gatech.edu} \\ \And
  Ming-Wei Chang \\
  Microsoft Research \\
  {\tt minchang@microsoft.com} \\}

\date{}

\begin{document}
\maketitle

\begin{figure}[h!]
\centering
\framebox {
  \parbox[l]{.9 \columnwidth}{ Original version: Proceedings of ACL
    2015, pp. 504-513. The experiments in the original version is done
    on the subset of the datasets. This revision of Sep 2016
    includes updated statistics for the actual datasets used in the
    original version and additional results for the full datasets (in appendix). While the experimental results are largely similar, we update the paper for the sake of
    completeness.} }
\end{figure}

\begin{abstract}

Non-linear models recently receive a lot of attention as people are starting to discover
the power of statistical and embedding
features. However, tree-based models are seldom studied in the context of structured learning despite their recent success on various classification and ranking tasks.
In this paper, we propose {\em \sysname}, a tree-based structured learning framework based on multiple additive regression trees. \sysname~is especially suitable for handling
tasks with dense features, and can be used to learn many different structures under various loss functions.

We apply \sysname~to the task of tweet entity linking --- a core component of tweet information extraction, which aims to identify and link name mentions to entities in a knowledge base. A novel inference algorithm is proposed to handle the special structure of the task.
The experimental results show that \sysname~significantly outperforms state-of-the-art tweet entity linking systems.

\end{abstract}

\section{Introduction}
\label{sec:intro}

Many natural language processing (NLP) problems can be formalized as structured prediction tasks. Standard algorithms for structured learning include Conditional Random Field (CRF)~\cite{lafferty2001conditional} and Structured Supported Vector Machine (SSVM)~\cite{tsochantaridis2004support}. These algorithms, usually equipped with a linear model and sparse lexical features, achieve state-of-the-art performances in many NLP applications such as part-of-speech tagging, named entity recognition and dependency parsing.


This classical combination of linear models and sparse features is challenged by the recent emerging usage of dense features such as statistical and embedding features.
Tasks with these low dimensional dense features require models to be more sophisticated to capture
the relationships between features. Therefore, non-linear models start to receive more attention as they are often more expressive than linear models.



Tree-based models such as boosted trees~\cite{friedman2001greedy} are flexible non-linear models. They can handle categorical features and count data better than other
non-linear models like Neural Networks. Unfortunately, to the best
 of our knowledge, little work has utilized tree-based methods for structured prediction, with the exception of TreeCRF~\cite{dietterich2004training}.



In this paper, we propose a novel structured learning framework called \sysname~(\textbf{S}tructured \textbf{M}ultiple \textbf{A}dditive \textbf{R}egression \textbf{T}rees).
Unlike TreeCRF, \sysname~is very versatile, as it can be applied to tasks beyond sequence tagging and can be trained under various objective functions. \sysname~is also powerful, as the high order relationships between features can be captured by non-linear regression trees.


We further demonstrate how \sysname~can be applied to tweet entity linking, an important and challenging task underlying many applications including product feedback~\cite{marketingcite1} and topic detection and tracking~\cite{mathioudakis2010twittermonitor}.
We apply \sysname~to entity linking using a simple logistic function as the loss function and propose a novel inference algorithm to prevent overlaps between entities. 


Our contributions are summarized as follows: 
\begin{itemize}
\item We propose a novel structured learning framework called \sysname. \sysname~combines non-linearity and efficiency of tree-based models with structured prediction, leading to a family of new algorithms. (Section~\ref{sec:model})
\item We apply \sysname~to tweet entity linking. Building on top of \sysname, we propose a novel inference algorithm for non-overlapping structure with the goal of preventing conflicting entity assignments. (Section~\ref{sec:inference})
\item We provide a systematic study of evaluation criteria in tweet entity linking by conducting extensive experiments over major data sets. The results show that \sysname~significantly outperforms state-of-the-art entity linking systems, including the system that is used to win the NEEL
     2014 challenge~\cite{microposts2014_neel_cano.ea:2014}. (Section~\ref{sec:exp}) 
\end{itemize}

\section{Structured Multiple Additive Regression Trees}
\label{sec:model}

The goal of a structured learning algorithm is to learn a joint scoring function $S$ between an input $\mb{x}$ and an output structure $\mb{y}$, $S:(\mb{x}, \mb{y}) \rightarrow \mathbb{R}$.
The structured output $\mb{y}$ often contains many interdependent variables, and the number of the possible structures can be
exponentially large with respect to the size of $\mb{x}$. At test time, the prediction $\mb{y}$ for $\mb{x}$ is obtained by
$$\argmax_{\mb{y}\in Gen(\mb{x})} S(\mb{x}, \mb{y}),$$
 where $Gen(\mb{x})$ represents the set of all valid output structures for $\mb{x}$.

Standard learning algorithms often directly optimize the model parameters. For example, assume that the joint scoring function $S$ is parameterized by $\theta$. Then, gradient descent algorithms can be used to optimize the model parameters $\theta$ iteratively. More specifically,
\begin{equation}
\label{eq:gd}
\theta_m = \theta_{m-1} - \eta_m \frac{\partial L(\mb{y}^*, S(\mb{x}, \mb{y}; \theta))}{\partial \theta_{m-1}} ,
\end{equation}
where $\mb{y}^*$ is the gold structure, $L(\mb{y}^*, S(\mb{x}, \mb{y}; \theta))$ is a loss function and $\eta_m$ is the learning rate of the $m$-th iteration.

In this paper, we propose a framework called {\bf Structured Multiple Additive Regression Trees (\sysname)}, which generalizes Multiple Additive Regression Trees (MART) for structured learning problems.
Different from Equation~(\ref{eq:gd}), \sysname~does {\em not} directly optimize the model parameters; instead, it approximates the optimal scoring function that minimize the loss by adding (weighted) regression tree models iteratively.

%
Due to the fact that there are exponentially many input-output pairs in the training data, $\sysname$ assumes that the joint scoring function can be decomposed as
\begin{equation*}
    S(\mb{x}, \mb{y}) = \sum_{k \in \Omega(\mb{x})} F(\mb{x}, \mb{y}_k),
\end{equation*}
where $\Omega(\mb{x})$ contains the set of the all factors for input $\mb{x}$ and $\mb{y}_k$ is the
sub-structure of $\mb{y}$ that corresponds to the $k$-th factor in $\Omega(\mb{x})$. For instance, in the task of word alignment, each factor can be defined as a pair of words from source and target languages respectively.
Note that we can recover
$\mb{y}$ from the union of $\{\mb{y}_k\}_1^K$.

The factor scoring function $F(\mb{x}, \mb{y}_k)$ can be optimized by performing gradient descent in the function space in the following manner:
\begin{equation}
\label{eq:fgd}
F_m(\mb{x}, \mb{y}_k) = F_{m-1}(\mb{x}, \mb{y}_k) - \eta_m g_m (\mb{x}, \mb{y}_k)
\end{equation}
where function $g_m (\mb{x}, \mb{y}_k)$
is the functional gradient.

Note that $g_m$ is a {\em function} rather than a vector. Therefore, modeling $g_m$ theoretically requires an infinite number of data points. We can address this difficulty by approximating $g_m$ with a finite number of point-wise functional gradients
\begin{align}
 & g_m  (\mb{x}, \mb{y}_k = u_k) = \label{eq:pwfg} \\
\notag & \left [ \frac{\partial L(\mb{y}^*, S(\mb{x}, \mb{y}_k = u_k))}{\partial F (\mb{x}, \mb{y}_k = u_k)} \right]_{F(\mb{x}, \mb{y}_k) = F_{m-1} (\mb{x}, \mb{y}_k)}
\end{align}
where $u_k$ index a valid sub-structure for the $k$-th factor of $\mb{x}$.

The key point of $\sysname$ is that it approximates $-g_m$ by modeling the point-wise negative functional gradients using a regression tree $h_m$. Then the factor scoring function can be obtained by
\begin{equation*}
F(\mb{x}, \mb{y}_k) = \sum_{m=1}^M \eta_m h_m (\mb{x}, \mb{y}_k),
\end{equation*}
where $h_m (\mb{x}, \mb{y}_k)$ is also called a basis function and $\eta_m$ can be simply set to 1~\cite{murphy2012machine}.


The detailed \sysname~algorithm is presented in Algorithm~\ref{algo:SMART}. The factor scoring function $F(\mb{x}, \mb{y}_k)$ is simply initialized to zero at first (line 1). After this, we iteratively update the function by adding regression trees. Note that the scoring function is shared by all the factors.
Specifically, given the current decision function $F_{m-1}$, we can consider
line 3 to line 9 a process of generating the pseudo training data $D$ for modeling the regression tree.
For each training example, \sysname~first computes the point-wise functional gradients according to Equation~(\ref{eq:pwfg}) (line 6). Here we use $g_{ku}$ as the abbreviation for $g_m (\mb{x}, \mb{y}_k = u_k)$.
In line 7, for each sub-structure $u_k$, we create a new training example for the regression problem by the feature vector $\Phi(\mb{x},\mb{y}_k = u_k)$ and the negative gradient $-g_{ku}$.
In line 10, a regression tree is constructed by minimizing differences between the prediction values and the point-wise negative gradients.
Then a new basis function (modeled by a regression tree)
will be added into the overall $F$ (line 11).

\begin{algorithm}[t]
\begin{small}
\caption{\sysname: A family of structured learning algorithms with multiple additive regression trees} \label{algo:SMART}
\begin{algorithmic}[1]
\BState $F_0 (\mb{x}, \mb{y}_k) =0$
\For{$m = 1 \text{ to } M$}: \Comment{going over all trees}
\State $D \leftarrow \emptyset$
\For{all examples}: \Comment{going over all examples}
\For{$\mb{y}_k\in\Omega(\mb{x})$}: \Comment{going over all factors}
\State \parbox[t]{\dimexpr\linewidth-\algorithmicindent}{For all $u_k$, obtain $g_{ku}$ by Equation~(\ref{eq:pwfg})}
\State  \parbox[t]{\dimexpr\linewidth-\algorithmicindent}{$D \leftarrow D \cup \{( \Phi(\mb{x},\mb{y}_k=u_k),-g_{ku}) \}$} 
\EndFor
\EndFor
\State \parbox[t]{\dimexpr\linewidth-\algorithmicindent}{$h_m (\mb{x}, \mb{y}_k) \leftarrow \texttt{TrainRegressionTree}(D)$ \strut}
\State $F_m (\mb{x}, \mb{y}_k) = F_{m-1} (\mb{x}, \mb{y}_k) + h_m (\mb{x}, \mb{y}_k)$
\EndFor
\end{algorithmic}
\end{small}
\end{algorithm}



It is crucial to note that \sysname~is {\em a family of algorithms} rather than a single algorithm. \sysname~is flexible
in the choice of the loss functions.
For example, we can use either logistic loss or hinge loss, which means that \sysname~can
train probabilistic models as well as non-probabilistic ones.
Depending on the choice of factors, \sysname~can handle various structures such as linear chains, trees, and even the semi-Markov chain~\cite{sarawagi2004semi}.

\paragraph{\sysname~versus MART}
There are two key differences between \sysname~and MART. First, \sysname~decomposes the joint scoring function $S(\mb{x}, \mb{y})$  into factors
to address the problem of the exploding number of input-output pairs for structured learning problems.
Second, \sysname~models a single scoring function $F(\mb{x}, \mb{y}_k)$ over inputs and output variables directly rather than $O$ different functions $F^o(\mb{x})$, each of which corresponds to a label class.
%
%
%
\paragraph{\sysname~versus TreeCRF}
TreeCRF can be viewed as a special case of~\sysname, and there are two points where \sysname~improves upon TreeCRF.
First, the model designed in~\cite{dietterich2004training} is tailored for
sequence tagging problems. Similar to MART, for a tagging task with $O$ tags, they choose to model $O$ functions $F^o(\mb{x}, o')$ instead
of directly modeling the joint score of the factor.  This imposes limitations on the feature functions, and TreeCRF is consequently unsuitable for many tasks such as entity linking.\footnote{For example, entity linking systems need to model the similarity between an entity and the document.
The TreeCRF formulation does not support such features.}
Second, \sysname~is more general in terms of the objective functions and applicable structures. In the next section, we will
see how \sysname~can be applied to a non-linear-chain structure and various loss functions. 

\section{\sysname~for Tweet Entity Linking}
\label{sec:inference}

\begin{figure*}[th!]
\centering
\includegraphics[scale=.48]{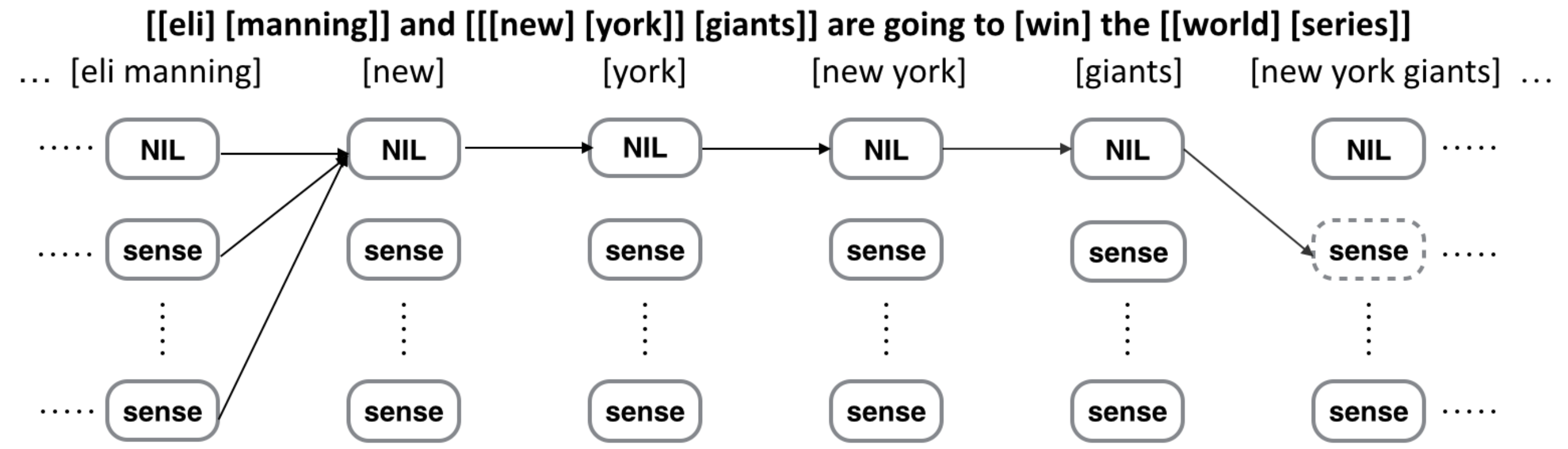}
\caption{\label{fig:forward} Example tweet and its mention candidates. Each mention candidate is marked as a pair of brackets in the original tweet and forms a column in the graph.  The graph demonstrates the non-overlapping constraint. To link the mention candidate ``new york giants'' to a non-\tb{Nil} entity, the system has to link previous four overlapping mention candidates to \textbf{Nil}. The mention candidate ``eli manning'' is not affected by ``new york giants''. {\bf Note that this is  not a standard linear chain problem.}}
\end{figure*}

We first formally define the task of tweet entity linking.
As \emph{input}, we are
given a tweet, an entity database
(\eg, Wikipedia where each article is an entity), and a lexicon\footnote{We use the standard techniques to construct the lexicon from anchor texts, redirect pages and other information resources.} which
maps a surface form into a set of entity candidates.
For each incoming tweet, all n-grams of this tweet will be used to find matches in the lexicon, and each match
will form a mention candidate.
As \emph{output},
we map every mention candidate (\eg,
``new york giants'') in the message to an entity (\eg, \textsc{New York Giants}) or to \textbf{Nil} (\ie, a non-entity).
A mention candidate can often potentially link to multiple entities, which we call possible {\em entity assignments}.

This task is a structured learning problem, as the final entity assignments of a tweet should not overlap with each other.\footnote{We follow the common practice and do not allow embedded entities.} We decompose this learning problem as follows: we make each mention candidate a factor, and the score of the
entity assignments of a tweet is the sum of the score of each entity and mention candidate pair. Although
all mention candidates are decomposed, the non-overlapping constraint requires the system to perform global inference.

Consider the example tweet in Figure~\ref{fig:forward}, where we show the tweet with the mention candidates in brackets.
To link the mention candidate ``new york giants'' to a non-\tb{Nil} entity, the system has to link previous overlapping mention candidates to \textbf{Nil}.
It is important to note that this is {\em \bf not} a linear chain problem because of the non-overlapping constraint, and  the inference algorithm needs to be carefully designed.

\subsection{Applying \sysname}

We derive specific model for tweet entity linking task with \sysname~and use logistic loss as our running example. The hinge loss version of the model can be derived in a similar way.

Note that the tweet and the mention candidates are given. Let $x$ be the tweet, $u_k$ be the entity assignment of
the $k$-th mention candidate. We use function
$F(\mb{x}, y_k = u_k)$ to model the score of the $k$-th mention candidate choosing entity $u_k$.\footnote{Note that each mention candidate has different
own entity sets.}
The overall scoring function can be decomposed as follows:
\begin{equation*}
S(\mb{x}, \mb{y} = \{u_k\}_{k=1}^K) = \sum_{k=1}^K F(\mb{x}, y_k=u_k)
\end{equation*}


\sysname~utilizes regression trees to model the scoring function $F(\mb{x}, y_k=u_k)$, which requires point-wise functional gradient for each entity of every mention candidate. Let's first write down the logistic loss function as
\begin{align}
\notag L (\mb{y}^*, S (\mb{x}, \mb{y})) =& -\log P(\mb{y}^* | \mb{x}) \\
\notag =& \log Z(\mb{x}) - S (\mb{x}, \mb{y}^*)
\end{align}
where $Z(\mb{x}) = \sum_{\mb{y}} \exp (S (\mb{x}, \mb{y}))$ is the potential function.
Then the point-wise gradients can be computed as
\begin{align*}
g_{ku} &= \frac{\partial L}{\partial F (\mb{x}, y_k=u_k)}  \\
&= P(y_k=u_k | \mb{x}) - \mb{1}[y^*_k=u_k],
\end{align*}
where $\mb{1}[\cdot]$ represents an indicator function. The conditional probability $P(y_k=u_k | \mb{x})$ can be computed by a variant of the forward-backward algorithm, which we will detail in the next subsection.




\subsection{Inference}
%
%
%
The non-overlapping structure is distinct from linear chain and semi-Markov chain~\cite{sarawagi2004semi} structures.
Hence, we propose a carefully designed forward-backward algorithm to calculate $P(y_k=u_k | \mb{x})$ based on current scoring function $F (\mb{x}, y_k=u_k)$ given by the regression trees. The non-overlapping constraint distinguishes our inference algorithm from
other forward-backward variants.

To compute the forward probability, we sort\footnote{Sorting helps the algorithms find non-overlapping candidates.} the mention candidates by their end indices and define forward recursion by
\begin{align}
\notag \alpha (u_1, 1) =& \exp (F (\mb{x}, y_1=u_1)) \\
\notag \alpha (u_k, k) =& \exp (F (\mb{x}, y_k=u_k)) \\
\notag & \cdot \prod_{p=1}^{P-1} \exp (F (\mb{x}, y_{k-p}=\textbf{Nil})) \\
& \cdot \sum_{u_{k-P}} \alpha (u_{k-P}, k-P) \label{eq:forward}
\end{align}
where $k-P$ is the index of the previous non-overlapping mention candidate. Intuitively, for the $k$-th mention candidate, we need to identify its nearest non-overlapping fellow and recursively compute the probability. The overlapping mention candidates can only take the \tb{Nil} entity.

Similarly, we can sort the mention candidates by their start indices and define backward recursion by
\begin{align}
\notag \beta(u_K, K) =& 1 \\
\notag \beta(u_k, k) =& \sum_{u_{k+Q}} \exp (F (\mb{x}, y_{k+Q}=u_{k+Q})) \\
\notag & \cdot \prod_{q=1}^{Q-1} \exp (F (\mb{x}, y_{k+q}=\textbf{Nil})) \\
& \cdot \beta (u_{k+Q}, k+Q) \label{eq:backward}
\end{align}
where $k+Q$ is the index of the next non-overlapping mention candidate. Note that the third terms of equation (\ref{eq:forward}) or (\ref{eq:backward}) will vanish if there are no corresponding non-overlapping mention candidates. 

Given the potential function can be computed by $Z(\mb{x}) = \sum_{u_k} \alpha(u_k, k) \beta(u_k, k)$,
%
%
for entities that are not \textbf{Nil},
\begin{align}
\notag P(y_k=u_k | \mb{x}) =& \frac{\exp (F (\mb{x}, y_k=u_k)) \cdot \beta(u_k, k)}{Z(\mb{x})} \\
\notag & \cdot \prod_{p=1}^{P-1} \exp (F (\mb{x}, y_{k-p}=\textbf{Nil})) \\
& \cdot \sum_{u_{k-P}} \alpha (u_{k-P}, k-P) \label{eq:func-grad}
\end{align}
The probability for the special token \textbf{Nil} can be obtained by
\begin{equation}
P(y_k=\textbf{Nil} | \mb{x}) = 1 - \sum_{u_k \neq \textbf{Nil}} P(y_k = u_k | \mb{x})
\end{equation}

In the worst case, the total cost of the forward-backward algorithm is $\mathcal{O}(\text{max}\{TK, K^2\})$, where $T$ is the number of entities of a mention candidate.\footnote{The cost is $\mathcal{O}(K^2)$ only if every mention candidate of the tweet overlaps other mention candidates. In practice, the algorithm is nearly linear w.r.t $K$.}

Finally, at test time, the decoding problem $\argmax_{\mb{y}} S(\mb{x}, \mb{y})$ can be solved by a variant of the Viterbi algorithm.

\subsection{Beyond \sysname: Modeling entity-entity relationships}
\label{sec:ent-ent}
It is important for entity linking systems to take advantage of the entity-to-entity information while making local decisions.
For instance, the identification of entity ``eli manning'' leads to a strong clue for linking ``new york giants'' to the NFL team.

Instead of defining a  more complicated structure and learning everything jointly, we employ a two-stage approach as the solution for modeling entity-entity relationships
after we found that~\sysname~achieves high precision and reasonable recall.
Specifically, in the first stage, the system identifies all possible entities with basic features, which enables the extraction of entity-entity features. In the second stage, we re-train \sysname~on a union of basic features and entity-entity features. We define entity-entity features based on the Jaccard distance introduced by~\newcite{guo2013link}.

Let $\Gamma (e_i)$ denotes the set of Wikipedia pages that contain a hyperlink to an entity $e_i$ and $\Gamma (t_{-i})$ denotes the set of pages that contain a hyperlink to any identified entity $e_j$ of the tweet $t$ in the first stage excluding $e_i$. The Jaccard distance between $e_i$ and $t$ is
\begin{equation*}
Jac(e_i, t) = \frac{|\Gamma (e_i) \cap \Gamma (t_{-i})|}{|\Gamma (e_i) \cup \Gamma (t_{-i})|} .
\end{equation*}
In addition to the Jaccard distance, we add one additional binary feature to indicate if the current entity has the highest Jaccard distance among all entities for this mention candidate.

\section{Experiments}
\label{sec:exp}


Our experiments are designed to answer the following three research questions in the context of tweet entity linking:
\begin{itemize}
  \item Do non-linear learning algorithms perform better than linear learning algorithms?
  \item Do structured entity linking models perform better than non-structured ones?
  \item How can we best capture the relationships between entities?
\end{itemize}


\subsection{Evaluation Methodology and Data}
We evaluate each entity linking system using two evaluation policies: Information Extraction (IE) driven evaluation and Information Retrieval (IR) driven evaluation.
For both evaluation settings, precision, recall and F1 scores are reported. Our data is constructed from two publicly available sources: Named Entity Extraction \& Linking (NEEL) Challenge~\cite{cano2014making} datasets, and the datasets released by~\newcite{fang2014entity}.
Note that we gather two datasets from ~\newcite{fang2014entity} and they
are used in two different evaluation settings. We refer to these two datasets as TACL-IE and TACL-IR, respectively.
We perform some data cleaning and unification on these sets.
The statistics of the datasets are presented in Table~\ref{tab:data}.\footnote{The datasets can be downloaded from \url{http://research.microsoft.com/en-us/downloads/24c267d7-4c19-41e8-8de1-2c116fcbdbd3/default.aspx}. We exclude Twitter messages that contain no ground truth entity mentions in the main evaluation. The statistics of the full datasets and the corresponding results are available in Appendix~\ref{app:results}.}


\paragraph{IE-driven evaluation}
The IE-driven evaluation is the standard evaluation for an end-to-end entity linking system. 
We follow~\newcite{carmel2014erd} and relax the definition of the correct mention boundaries, as they are often ambiguous.
A mention boundary is considered to be correct if it overlaps (instead of being the same) with the gold mention boundary. Please see~\cite{carmel2014erd} for more details on the procedure of calculating the precision, recall and F1 score.
%

The NEEL and TACL-IE datasets have different annotation guidelines and different choices of knowledge bases, so we perform the following procedure to clean the data and unify the annotations. We first filter out the annotations that link to entities excluded by our knowledge base. We use the same knowledge base as the ERD 2014 competition~\cite{carmel2014erd}, which includes the union of entities in Wikipedia and Freebase.  Second, we follow NEEL annotation guideline and re-annotate TACL-IE dataset. For instance, in order to be consistent with NEEL, all the user tags (e.g. @BarackObama) are re-labeled as entities in TACL-IE.

We train all the models with NEEL Train dataset and evaluate different systems on NEEL Test and TACL-IE datasets.
In addition, we sample 800 tweets from NEEL Train dataset as our development set to perform parameter tuning.


\paragraph{IR-driven evaluation}
The IR-driven evaluation is proposed by ~\newcite{fang2014entity}. It is motivated by a key application of entity linking --- retrieval of relevant tweets for target entities, which is crucial for downstream applications such as product research and sentiment analysis.
In particular, given a query entity we can search for tweets based on the match with some potential surface forms of the query entity. Then, an entity linking system is evaluated by its ability to correctly identify the presence or absence of the query entity in every tweet.
%
Our IR-driven evaluation is based on the TACL-IR set, which includes 980 tweets sampled for ten query entities of five entity types (roughly 100 tweets per entity). About 37\% of the sampled tweets did not mention the query entity due to the anchor ambiguity.


\begin{table} [t]
\centering
\small\addtolength{\tabcolsep}{-2pt}
\begin{tabular}{ llll}
    \hline
    Data & \#Tweet & \#Entity & Date \\ \hline
    NEEL Train & 1171 & 2202 & Jul. \textasciitilde Aug. 11\\
    NEEL Test & 398 & 687 & Jul. \textasciitilde Aug. 11 \\
    TACL-IE & 180 & 300 & Dec. 12 \\
    TACL-IR & 980 & NA & Dec. 12 \\
    \hline
\end{tabular}
\caption{Statistics of data sets.}
\label{tab:data}
\end{table}

\subsection{Experimental Settings}
\label{sec:exp-setting}

\paragraph{Features}
We employ a total number of 37 dense features as our basic feature set. Most of the features are adopted from~\cite{guo2013link}\footnote{We consider features of Base, Capitalization Rate, Popularity, Context Capitalization and Entity Type categories.}, including
various statistical features such as the probability of the surface to be used as anchor text in Wikipedia.
We also add additional Entity Type features correspond to the following entity types: Character, Event, Product and Brand. Finally, we include several NER features to indicate each mention candidate belongs to one the following NER types: Twitter user, Twitter hashtag, Person, Location, Organization, Product, Event and Date.


%
%


\paragraph{Algorithms}

Table~\ref{tab:models} summarizes all the algorithms  that are compared in our experiments. First, we consider two linear structured learning algorithms: Structured Perceptron~\cite{collins2002discriminative} and Linear Structured SVM (SSVM)~\cite{tsochantaridis2004support}. 

For non-linear models, we consider polynomial SSVM, which employs polynomial kernel inside the structured SVM algorithm. We also include LambdaRank~\cite{quoc2007learning},
a neural-based learning to rank algorithm, which is widely used in the information retrieval literature.
We further compare with MART, which is designed for performing multiclass classification using log loss without considering the structured information. Finally, we have our proposed log-loss \sysname~algorithm, as described in Section~\ref{sec:inference}.
\footnote{Our pilot experiments show that the log-loss \sysname~consistently outperforms the hinge-loss \sysname.}

Note that our baseline systems are quite strong. Linear SSVM has been used in one of the state-of-the-art tweet entity linking systems~\cite{guo2013link}, and the system based on MART is the winning system of the 2014 NEEL Challenge~\cite{microposts2014_neel_cano.ea:2014}\footnote{Note that the numbers we reported here are different from the results in NEEL challenge due to the fact that we have cleaned the datasets and the evaluation metrics are slightly different in this paper.}.

 Table~\ref{tab:models} summarizes several properties of the algorithms. For example, most algorithms are structured (e.g. they perform dynamic programming at test time) except for MART and LambdaRank, which treat mention candidates independently.

%

\begin{table} [t]
\centering
\small\addtolength{\tabcolsep}{-2pt}
\begin{tabular}{ llll}
    \hline
    Model & Structured & Non-linear & Tree-based \\ \hline
    Structured Perceptron & \checkmark  & & \\
    Linear SSVM & \checkmark &  & \\
    Polynomial SSVM   & \checkmark & \checkmark & \\
    LambdaRank & &\checkmark & \\
    MART        &  & \checkmark & \checkmark \\
    \sysname    & \checkmark & \checkmark & \checkmark \\
    \hline
\end{tabular}
\caption{Included algorithms and their properties.}
\label{tab:models}
\end{table}

\paragraph{Parameter tuning}
All the hyper-parameters are tuned on the development set. Then, we re-train our models on full training data (including the dev set) with the best parameters. We choose the soft margin parameter $C$ from $\{0.5, 1, 5, 10 \}$ for two structured SVM methods. After a preliminary parameter search, we fixed the number of trees to 300 and the minimum number of documents in a leaf to 30 for all tree-based models. 
For LambdaRank, we use a two layer feed forward network. We select the number of hidden units from $\{10,20,30,40\}$ and learning rate from $\{0.1,0.01,0.001\}$.

It is widely known that F1 score can be affected by the trade-off between precision and recall. In order to make the comparisons between
all algorithms fairer in terms of F1 score, we include a post-processing step to balance precision and recall for all the systems. Note the tuning is only conducted for the purpose of robust evaluation. In particular, we adopt a simple tuning strategy that works well for all the algorithms, in which
we add a bias term $b$ to the scoring function value of \tb{Nil}:
$$F(\mb{x}, y_k =\tb{Nil}) \leftarrow F(\mb{x}, y_k =\tb{Nil}) + b .$$
We choose the bias term $b$ from values between $-3.0$ to $3.0$ on the dev set and apply the same bias term at test time.

\subsection{Results}

Table~\ref{tab:results} presents the empirical findings for \sysname~and competitive methods on tweet entity linking task in both IE and IR settings. In the following,
we analyze the empirical results in details.

\begin{table*} [ht!]
\centering
\addtolength{\tabcolsep}{-2pt}
\begin{tabular}{ l | ccc | ccc | ccc | ccc }
    \hline
    \multirow{2}{*}{Model} & \multicolumn{3}{c|}{NEEL Dev} & \multicolumn{3}{c|}{NEEL Test} & \multicolumn{3}{c|}{TACL-IE} & \multicolumn{3}{c}{TACL-IR}                                \\
                        & P & R & F1 & P & R & F1 & P & R & F1 & P & R & F1 \\ \hline
  Structured Perceptron & 75.8 & 62.8 & 68.7 & 79.1 & 64.3 & 70.9 & 74.4 & 63.0 & 68.2 & 86.2 & 43.8 & 58.0 \\
  Linear SSVM           & 78.0 & 66.1 & 71.5 & 80.5 & 67.1 & 73.2 & \tb{78.2} & 64.7 & 70.8 & 86.7 & 48.5 & 62.2 \\ \hline
  Polynomial SSVM       & 77.7 & 70.7 & 74.0 & 81.3 & 69.0 & 74.6 & 76.8 & 64.0 & 69.8 & 91.1 & 48.8 & 63.6 \\
  LambdaRank            & 75.0 & 69.0 & 71.9 & 80.3 & 71.2 & 75.5 & 77.8 & 66.7   & 71.8   & 85.8   & 42.4  & 56.8 \\
  MART                  & 76.2 & 74.3 & 75.2 & 76.8 & 78.0 & 77.4 & 73.4 & 71.0 & 72.2 & \tb{98.1} & 46.4 & 63.0 \\
  \sysname              & \tb{79.1} & \tb{75.8} & \tb{77.4} & \tb{83.2} & \tb{79.2} & \tb{81.1} & 76.8 & \tb{73.0} & \tb{74.9} & 95.1 & \tb{52.2} & \tb{67.4} \\ \hline
  + entity-entity      & \underline{79.2} & \underline{75.8} & \underline{77.5} & 81.5 & 76.4 & 78.9 & 77.3 & \underline{73.7} & \underline{75.4} & 95.5 & \underline{56.7} & \underline{71.1} \\
    \hline
\end{tabular}
\caption{IE-driven and IR-driven evaluation results for different models. The best results with basic features are in \tb{bold}. The results are \underline{underlined} if adding entity-entity features gives the overall best results. Please see Appendix.}
\label{tab:results}
\end{table*}

\begin{figure}[t]
\centering
\begin{tikzpicture}[scale=1]
      \begin{axis}
        [
        cycle list name=mylist,
        ylabel= F1 score, 
        ylabel near ticks,
        xlabel = Bias,
        xlabel near ticks,
        label style={font=\footnotesize},
        legend style={font=\tiny},
        ymin = 50, ymax=80,
        xmax = 2, xmin =-2,
        mark size = 1,
        x tick label style={anchor=north east, inner sep=1mm},
        width=\textwidth*0.5,
        height= 6cm,
        legend pos=south west,
        every axis/.append style={font=\tiny},
        mark size = 2
        ]

        \addplot coordinates
        {
          (3.0,68.03)
          (2.0,68.48)
          (1.5,68.48)
          (1.0,68.47)
          (0.5,68.57)
          (0.25,68.67)
          (0.1,68.67)
          (0.0,68.67)
          (-0.1,68.67)
          (-0.25,68.67)
          (-0.5,68.67)
          (-1.0,68.56)
          (-1.5,68.51)
          (-2.0,68.41)
          (-3.0,68.15)
        };

        \addplot coordinates
        {
          (3.0,0.27)
          (2.0,7.66)
          (1.5,23.13)
          (1.0,41.70)
          (0.5,60.67)
          (0.25,65.88)
          (0.1,68.39)
          (0.0,69.08)
          (-0.1,69.97)
          (-0.25,71.52)
          (-0.5,71.06)
          (-1.0,66.71)
          (-1.5,56.12)
          (-2.0,38.08)
          (-3.0,17.58)
        };

        \addplot coordinates
        {
          (3.0,14.88)
          (2.0,30.24)
          (1.5,41.85)
          (1.0,58.51)
          (0.5,69.14)
          (0.25,71.65)
          (0.1,72.62)
          (0.0,73.23)
          (-0.1,74.05)
          (-0.25,73.79)
          (-0.5,72.30)
          (-1.0,66.27)
          (-1.5,52.17)
          (-2.0,38.38)
          (-3.0,19.60)
        };

        \addplot coordinates
        {
          (3.0,35.54)
          (2.0,54.72)
          (1.5,61.46)
          (1.0,65.65)
          (0.5,69.95)
          (0.25,71.55)
          (0.1,71.66)
          (0.0,72.09)
          (-0.1,72.52)
          (-0.25,73.94)
          (-0.5,74.68)
          (-1.0,74.68)
          (-1.5,75.24)
          (-2.0,74.29)
          (-3.0,70.21)
        };
        \addplot coordinates
        {
          (3.0,26.45)
          (2.0,41.28)
          (1.5,50.17)
          (1.0,59.15)
          (0.5,67.39)
          (0.25,70.71)
          (0.1,71.91)
          (0.0,71.79)
          (-0.1,70.92)
          (-0.25,69.40)
          (-0.5,64.43)
          (-1.0,49.04)                              
        };
        
        \addplot coordinates
        {
          (3.0,40.72)
          (2.0,59.74)
          (1.5,66.89)
          (1.0,71.67)
          (0.5,74.83)
          (0.25,75.42)
          (0.1,75.83)
          (0.0,75.73)
          (-0.1,75.96)
          (-0.25,76.20)
          (-0.5,76.56)
          (-1.0,77.01)
          (-1.5,77.42)
          (-2.0,76.42)
          (-3.0,73.49)
        };

        \legend{SP,Linear SSVM,Poly. SSVM,MART,NN,\sysname};
      \end{axis}
\end{tikzpicture}
\caption{\label{fig:nil} Balance precisions and recalls. X-axis corresponds to values of the bias terms for the special token \tb{Nil}. Note that \sysname~is still the overall winning system without tuning the threshold.}
\end{figure}
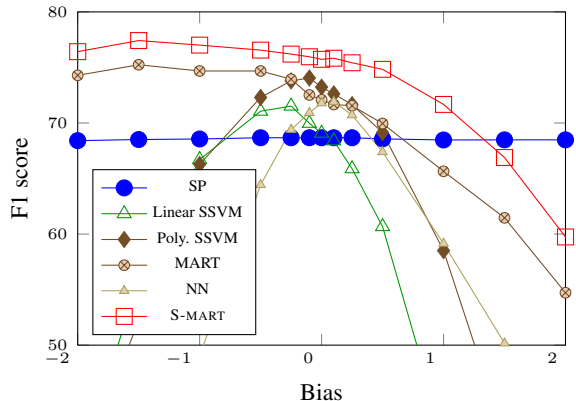

\paragraph{Linear models vs. non-linear models}
Table~\ref{tab:results} clearly shows that linear models perform worse than non-linear models when they are restricted to the IE setting of the tweet entity linking task. The story is similar in IR-driven evaluation, with the exception of LambdaRank. Among the linear models, linear SSVM demonstrates its superiority over Structured Perceptron on all datasets, which aligns with the results of~\cite{tsochantaridis2005large} on the named entity recognition task.

We have many interesting observations on the non-linear models side. First, by adopting a polynomial kernel, the non-linear SSVM further improves the entity linking performances on the NEEL datasets and TACL-IR dataset. Second, LambdaRank, a neural network based model, achieves better results than linear models in IE-driven evaluation, but the results in IR-driven evaluation are worse than all the other methods. We believe the reason for this dismal performance is that the neural-based method tends to overfit the IR setting given the small number of training examples.
Third, both MART and \sysname~significantly outperform alternative linear and non-linear methods in IE-driven evaluation and performs better or similar to other methods in IR-driven evaluation. This suggests that tree-based non-linear models are suitable for tweet entity linking task. Finally,  \sysname~outperforms previous state-of-the-art method Structured SVM by a surprisingly large margin. In the NEEL Test dataset, the difference is more than 10\% F1.
Overall, the results show that the shallow linear models are not expressive enough to capture the complex patterns in the data, which are represented by a few dense features.



\paragraph{Structured learning models}

To showcase structured learning technique is crucial for entity linking with non-linear models, we compare \sysname~against MART directly.
As shown in Table~\ref{tab:results}, \sysname~can achieve higher precision and recall points compared to MART on all datasets in terms of IE-driven evaluation, and can improve F1 by 4 points on NEEL Test and TACL-IR datasets. The task of entity linking is to produce non-overlapping entity assignments that match the gold mentions. By adopting structured learning technique, \sysname~is able to automatically take into account the non-overlapping constraint during learning and inference, and produce global optimal entity assignments for mention candidates of a tweet. One effect is that \sysname~can easily eliminate some common errors caused by popular entities (e.g. new york in Figure~\ref{fig:forward}).

\paragraph{Modeling entity-entity relationships}

Entity-entity relationships provide strong clues for entity disambiguation. In this paper, we use the simple two-stage approach described in Section~\ref{sec:ent-ent} to capture the relationships between entities.
As shown in Table~\ref{tab:results}, the significant improvement in IR-driven evaluation indicates the importance of incorporating entity-entity information.

Interestingly, while IR-driven results are significantly improved, IE-driven results are similar or even worse given entity-entity features. We believe the reason is that IE-driven and IR-driven evaluations focus on different aspects of tweet entity linking task.
As~\newcite{guo2013link} shows that most mentions in tweets should be linked to the most popular entities, IE setting actually pays more attention on mention detection sub-problem.
In contrast to IE setting, IR setting focuses on entity disambiguation, since we only need to decide whether the tweet is relevant to the query entity.
Therefore, we believe that both evaluation policies are needed for tweet entity linking.

\paragraph{Balance Precision and Recall}
Figure~\ref{fig:nil} shows the results of tuning the bias term for balancing precision and recall on the dev set. The results show that~\sysname~outperforms competitive approaches without any tuning, with similar margins to the results after tuning. Balancing precision and recall improves F1 scores for all the systems, which suggests that the simple tuning method performs quite well. Finally, we have an interesting observation that different methods have various scales of model scores.




\section{Related Work}
\label{sec:related}

Linear structured learning methods have been proposed and widely used in the literature. Popular models include Structured Perceptron~\cite{collins2002discriminative}, Conditional Random Field~\cite{lafferty2001conditional} and Structured SVM~\cite{roller2004max,tsochantaridis2005large}. 
Recently, many structured learning models based on neural networks have been proposed and are widely used in language modeling~\cite{bengio2006neural,mikolov2010recurrent}, sentiment classification~\cite{socher2013recursive}, as well as parsing~\cite{socher2011parsing}.~\newcite{cortes2014learning} recently proposed a boosting framework which treats different structured learning algorithms as base learners to ensemble structured prediction results.

 Tree-based models have been shown to provide more robust and accurate performances than neural networks in some tasks of computer vision~\cite{roe2005boosted,babenko2011robust} and information retrieval~\cite{li2007mcrank,wu2010adapting}, suggesting that it is worth to investigate tree-based non-linear models for structured learning problems. To the best of our knowledge, TreeCRF~\cite{dietterich2004training} is the only work that explores tree-based methods for structured learning problems. The relationships between TreeCRF and our work
have been discussed in Section~\ref{sec:model}.\footnote{\newcite{chen2015efficient} independently propose a class of tree-based structured learning models using a similar formalization of \sysname. However, they focus on exploring second-order information of line chain structures, while we aim at handling different structures and objective functions.}

Early research on entity linking has focused on well written documents
\cite{bunescu2006using,cucerzan2007large,MilneWi08}. Due to the raise of social media, many techniques
have been proposed or tailored to short texts including
tweets, for the problem of entity linking \cite{FerraginaSc10,MeijWeRi12,guo2013link} as well as the related problem
of named entity recognition (NER) \cite{RCME11}. Recently, non-textual information such
as spatial and temporal signals have also been used to improve entity linking systems~\cite{fang2014entity}.
The task of entity linking has attracted a lot of attention, and many shared tasks have been hosted
to promote entity linking research~\cite{ji2010overview,ji2011knowledge,microposts2014_neel_cano.ea:2014,carmel2014erd}.

Building an end-to-end entity linking system involves in solving two interrelated sub-problems: mention detection and entity disambiguation. Earlier research on entity linking has been largely focused on the entity disambiguation problem, including most work on entity linking for well-written documents such as news and encyclopedia articles~\cite{cucerzan2007large} and also few for tweets~\cite{liu2013entity}. Recently,
people have focused on building systems that consider mention detection and entity disambiguation jointly.
For example, 
 ~\newcite{cucerzan2012msr} delays the mention detection decision and consider the mention detection and entity linking problem jointly. Similarly,
 ~\newcite{sil2013re} proposed to use a reranking approach to obtain overall better results on mention detection and entity disambiguation.


\section{Conclusion and Future Work}
\label{sec:con}
In this paper, we propose \sysname, {a family of structured learning algorithms} which is flexible on the choices of the loss functions and structures.
We demonstrate the power of \sysname~by applying it to tweet entity linking, and it significantly outperforms the current state-of-the-art entity linking systems.
In the future, we would like to investigate the advantages and disadvantages between tree-based models and other non-linear models such as deep neural networks or recurrent neural networks.

\paragraph{Acknowledgments} We thank the reviewers for their insightful feedback. We also thank Yin Li and Ana Smith for their valuable comments on earlier version of this paper.


\bibliographystyle{acl}
\bibliography{ref}

\begin{thebibliography}{}

\bibitem[\protect\citename{Asur and Huberman}2010]{marketingcite1}
S.~Asur and B.A. Huberman.
\newblock 2010.
\newblock Predicting the future with social media.
\newblock {\em arXiv preprint arXiv:1003.5699}.

\bibitem[\protect\citename{Babenko \bgroup et al.\egroup
  }2011]{babenko2011robust}
Boris Babenko, Ming-Hsuan Yang, and Serge Belongie.
\newblock 2011.
\newblock Robust object tracking with online multiple instance learning.
\newblock {\em Pattern Analysis and Machine Intelligence, IEEE Transactions
  on}, pages 1619--1632.

\bibitem[\protect\citename{Bengio \bgroup et al.\egroup
  }2006]{bengio2006neural}
Yoshua Bengio, Holger Schwenk, Jean-S{\'e}bastien Sen{\'e}cal, Fr{\'e}deric
  Morin, and Jean-Luc Gauvain.
\newblock 2006.
\newblock Neural probabilistic language models.
\newblock In {\em Innovations in Machine Learning}, pages 137--186.

\bibitem[\protect\citename{Bunescu and Pasca}2006]{bunescu2006using}
R.~C Bunescu and M.~Pasca.
\newblock 2006.
\newblock Using encyclopedic knowledge for named entity disambiguation.
\newblock In {\em Proceedings of the European Chapter of the ACL (EACL)}, pages
  9--16.

\bibitem[\protect\citename{Burges \bgroup et al.\egroup
  }2007]{quoc2007learning}
Christopher~JC Burges, Robert Ragno, and Quoc Le, Qu.
\newblock 2007.
\newblock Learning to rank with nonsmooth cost functions.
\newblock In {\em Advances in neural information processing systems (NIPS)},
  pages 193--200.

\bibitem[\protect\citename{Cano and
  others}2014]{microposts2014_neel_cano.ea:2014}
AE~Cano et~al.
\newblock 2014.
\newblock Microposts2014 neel challenge.
\newblock In {\em Microposts2014 NEEL Challenge}.

\bibitem[\protect\citename{Cano \bgroup et al.\egroup }2014]{cano2014making}
Amparo~E Cano, Giuseppe Rizzo, Andrea Varga, Matthew Rowe, Milan Stankovic, and
  Aba-Sah Dadzie.
\newblock 2014.
\newblock Making sense of microposts (\# microposts2014) named entity
  extraction \& linking challenge.
\newblock {\em Making Sense of Microposts (\# Microposts2014)}.

\bibitem[\protect\citename{Carmel \bgroup et al.\egroup }2014]{carmel2014erd}
David Carmel, Ming-Wei Chang, Evgeniy Gabrilovich, Bo-June~Paul Hsu, and
  Kuansan Wang.
\newblock 2014.
\newblock Erd'14: entity recognition and disambiguation challenge.
\newblock In {\em ACM SIGIR Forum}, pages 63--77.

\bibitem[\protect\citename{Chen \bgroup et al.\egroup }2015]{chen2015efficient}
Tianqi Chen, Sameer Singh, Ben Taskar, and Carlos Guestrin.
\newblock 2015.
\newblock Efficient second-order gradient boosting for conditional random
  fields.
\newblock In {\em Proceedings of the International Conference on Artificial
  Intelligence and Statistics}, pages 147--155.

\bibitem[\protect\citename{Collins}2002]{collins2002discriminative}
Michael Collins.
\newblock 2002.
\newblock Discriminative training methods for hidden markov models: Theory and
  experiments with perceptron algorithms.
\newblock In {\em Proceedings of the conference on Empirical methods in natural
  language processing (EMNLP)}, pages 1--8.

\bibitem[\protect\citename{Cortes \bgroup et al.\egroup
  }2014]{cortes2014learning}
Corinna Cortes, Vitaly Kuznetsov, and Mehryar Mohri.
\newblock 2014.
\newblock Learning ensembles of structured prediction rules.
\newblock In {\em Proceedings of the Annual Meeting of the Association for
  Computational Linguistics (ACL)}.

\bibitem[\protect\citename{Cucerzan}2007]{cucerzan2007large}
Silviu Cucerzan.
\newblock 2007.
\newblock Large-scale named entity disambiguation based on wikipedia data.
\newblock In {\em Proceedings of the Conference on Empirical Methods in Natural
  Language Processing (EMNLP)}, pages 708--716.

\bibitem[\protect\citename{Cucerzan}2012]{cucerzan2012msr}
Silviu Cucerzan.
\newblock 2012.
\newblock The msr system for entity linking at tac 2012.
\newblock In {\em Text Analysis Conference}.

\bibitem[\protect\citename{Dietterich \bgroup et al.\egroup
  }2004]{dietterich2004training}
Thomas~G Dietterich, Adam Ashenfelter, and Yaroslav Bulatov.
\newblock 2004.
\newblock Training conditional random fields via gradient tree boosting.
\newblock In {\em Proceedings of the International Conference on Machine
  Learning (ICML)}, pages 28--35.

\bibitem[\protect\citename{Fang and Chang}2014]{fang2014entity}
Yuan Fang and Ming-Wei Chang.
\newblock 2014.
\newblock Entity linking on microblogs with spatial and temporal signals.
\newblock {\em Transactions of the Association for Computational Linguistics
  (ACL)}, pages 259--272.

\bibitem[\protect\citename{Ferragina and Scaiella}2010]{FerraginaSc10}
P.~Ferragina and U.~Scaiella.
\newblock 2010.
\newblock {TAGME}: on-the-fly annotation of short text fragments (by
  {Wikipedia} entities).
\newblock In {\em Proceedings of ACM Conference on Information and Knowledge
  Management (CIKM)}, pages 1625--1628.

\bibitem[\protect\citename{Friedman}2001]{friedman2001greedy}
Jerome~H Friedman.
\newblock 2001.
\newblock Greedy function approximation: a gradient boosting machine.
\newblock {\em Annals of Statistics}, pages 1189--1232.

\bibitem[\protect\citename{Guo \bgroup et al.\egroup }2013]{guo2013link}
Stephen Guo, Ming-Wei Chang, and Emre Kiciman.
\newblock 2013.
\newblock To link or not to link? a study on end-to-end tweet entity linking.
\newblock In {\em Proceedings of the North American Chapter of the Association
  for Computational Linguistics (NAACL)}, pages 1020--1030.

\bibitem[\protect\citename{Ji and Grishman}2011]{ji2011knowledge}
Heng Ji and Ralph Grishman.
\newblock 2011.
\newblock Knowledge base population: Successful approaches and challenges.
\newblock In {\em Proceedings of the Annual Meeting of the Association for
  Computational Linguistics (ACL)}, pages 1148--1158.

\bibitem[\protect\citename{Ji \bgroup et al.\egroup }2010]{ji2010overview}
Heng Ji, Ralph Grishman, Hoa~Trang Dang, Kira Griffitt, and Joe Ellis.
\newblock 2010.
\newblock Overview of the tac 2010 knowledge base population track.
\newblock In {\em Third Text Analysis Conference (TAC)}.

\bibitem[\protect\citename{Lafferty \bgroup et al.\egroup
  }2001]{lafferty2001conditional}
John Lafferty, Andrew McCallum, and Fernando~CN Pereira.
\newblock 2001.
\newblock Conditional random fields: Probabilistic models for segmenting and
  labeling sequence data.
\newblock In {\em Proceedings of the International Conference on Machine
  Learning (ICML)}, pages 282--289.

\bibitem[\protect\citename{Li \bgroup et al.\egroup }2007]{li2007mcrank}
Ping Li, Qiang Wu, and Christopher~J Burges.
\newblock 2007.
\newblock Mcrank: Learning to rank using multiple classification and gradient
  boosting.
\newblock In {\em Advances in neural information processing systems (NIPS)},
  pages 897--904.

\bibitem[\protect\citename{Liu \bgroup et al.\egroup }2013]{liu2013entity}
Xiaohua Liu, Yitong Li, Haocheng Wu, Ming Zhou, Furu Wei, and Yi~Lu.
\newblock 2013.
\newblock Entity linking for tweets.
\newblock In {\em Proceedings of the Association for Computational Linguistics
  (ACL)}, pages 1304--1311.

\bibitem[\protect\citename{Mathioudakis and
  Koudas}2010]{mathioudakis2010twittermonitor}
Michael Mathioudakis and Nick Koudas.
\newblock 2010.
\newblock Twittermonitor: trend detection over the twitter stream.
\newblock In {\em Proceedings of the ACM SIGMOD International Conference on
  Management of data (SIGMOD)}, pages 1155--1158.

\bibitem[\protect\citename{Meij \bgroup et al.\egroup }2012]{MeijWeRi12}
E.~Meij, W.~Weerkamp, and M.~de~Rijke.
\newblock 2012.
\newblock Adding semantics to microblog posts.
\newblock In {\em Proceedings of International Conference on Web Search and Web
  Data Mining (WSDM)}, pages 563--572.

\bibitem[\protect\citename{Mikolov \bgroup et al.\egroup
  }2010]{mikolov2010recurrent}
Tomas Mikolov, Martin Karafi{\'a}t, Lukas Burget, Jan Cernock{\`y}, and Sanjeev
  Khudanpur.
\newblock 2010.
\newblock Recurrent neural network based language model.
\newblock In {\em INTERSPEECH}, pages 1045--1048.

\bibitem[\protect\citename{Milne and Witten}2008]{MilneWi08}
D.~Milne and I.~H. Witten.
\newblock 2008.
\newblock Learning to link with {Wikipedia}.
\newblock In {\em Proceedings of ACM Conference on Information and Knowledge
  Management (CIKM)}, pages 509--518.

\bibitem[\protect\citename{Murphy}2012]{murphy2012machine}
Kevin~P Murphy.
\newblock 2012.
\newblock {\em Machine learning: a probabilistic perspective}.
\newblock MIT press.

\bibitem[\protect\citename{Ritter \bgroup et al.\egroup }2011]{RCME11}
A.~Ritter, S.~Clark, Mausam, and O.~Etzioni.
\newblock 2011.
\newblock Named entity recognition in tweets: an experimental study.
\newblock In {\em Proceedings of the Conference on Empirical Methods for
  Natural Language Processing (EMNLP)}, pages 1524--1534.

\bibitem[\protect\citename{Roe \bgroup et al.\egroup }2005]{roe2005boosted}
Byron~P Roe, Hai-Jun Yang, Ji~Zhu, Yong Liu, Ion Stancu, and Gordon McGregor.
\newblock 2005.
\newblock Boosted decision trees as an alternative to artificial neural
  networks for particle identification.
\newblock {\em Nuclear Instruments and Methods in Physics Research Section A:
  Accelerators, Spectrometers, Detectors and Associated Equipment}, pages
  577--584.

\bibitem[\protect\citename{Sarawagi and Cohen}2004]{sarawagi2004semi}
Sunita Sarawagi and William~W Cohen.
\newblock 2004.
\newblock Semi-markov conditional random fields for information extraction.
\newblock In {\em Advances in Neural Information Processing Systems (NIPS)},
  pages 1185--1192.

\bibitem[\protect\citename{Sil and Yates}2013]{sil2013re}
Avirup Sil and Alexander Yates.
\newblock 2013.
\newblock Re-ranking for joint named-entity recognition and linking.
\newblock In {\em Proceedings of ACM Conference on Information and Knowledge
  Management (CIKM)}, pages 2369--2374.

\bibitem[\protect\citename{Socher \bgroup et al.\egroup
  }2011]{socher2011parsing}
Richard Socher, Cliff~C Lin, Chris Manning, and Andrew~Y Ng.
\newblock 2011.
\newblock Parsing natural scenes and natural language with recursive neural
  networks.
\newblock In {\em Proceedings of the International Conference on Machine
  Learning (ICML)}, pages 129--136.

\bibitem[\protect\citename{Socher \bgroup et al.\egroup
  }2013]{socher2013recursive}
Richard Socher, Alex Perelygin, Jean~Y Wu, Jason Chuang, Christopher~D Manning,
  Andrew~Y Ng, and Christopher Potts.
\newblock 2013.
\newblock Recursive deep models for semantic compositionality over a sentiment
  treebank.
\newblock In {\em Proceedings of the Conference on Empirical Methods in Natural
  Language Processing (EMNLP)}, pages 1631--1642.

\bibitem[\protect\citename{Taskar \bgroup et al.\egroup }2004]{roller2004max}
Ben Taskar, Carlos Guestrin, and Daphne Roller.
\newblock 2004.
\newblock Max-margin markov networks.
\newblock In {\em Advances in Neural Information Processing Systems (NIPS)}.

\bibitem[\protect\citename{Tsochantaridis \bgroup et al.\egroup
  }2004]{tsochantaridis2004support}
Ioannis Tsochantaridis, Thomas Hofmann, Thorsten Joachims, and Yasemin Altun.
\newblock 2004.
\newblock Support vector machine learning for interdependent and structured
  output spaces.
\newblock In {\em Proceedings of the International Conference on Machine
  Learning (ICML)}, page 104.

\bibitem[\protect\citename{Tsochantaridis \bgroup et al.\egroup
  }2005]{tsochantaridis2005large}
Ioannis Tsochantaridis, Thorsten Joachims, Thomas Hofmann, and Yasemin Altun.
\newblock 2005.
\newblock Large margin methods for structured and interdependent output
  variables.
\newblock In {\em Journal of Machine Learning Research}, pages 1453--1484.

\bibitem[\protect\citename{Wu \bgroup et al.\egroup }2010]{wu2010adapting}
Qiang Wu, Christopher~JC Burges, Krysta~M Svore, and Jianfeng Gao.
\newblock 2010.
\newblock Adapting boosting for information retrieval measures.
\newblock {\em Information Retrieval}, pages 254--270.

\end{thebibliography}

\clearpage
\onecolumn
\begin{appendices}
\section{Appendix: Results of Full Datasets}
\label{app:results}

\begin{table} [h!]
\centering
\addtolength{\tabcolsep}{-2pt}
\begin{tabular}{ llll}
    \hline
    Data & \#Tweet & \#Entity & Date \\ \hline
    NEEL Train & 2340 & 2202 & Jul. \textasciitilde Aug. 11\\
    NEEL Test & 1164 & 687 & Jul. \textasciitilde Aug. 11 \\
    TACL-IE & 500 & 300 & Dec. 12 \\
    \hline
\end{tabular}
\caption{Statistics of data sets introduced in Section~\ref{sec:exp-setting}.}
\label{tab:full-data}
\end{table}

\begin{table*} [ht!]
\centering
\addtolength{\tabcolsep}{-2pt}
\begin{tabular}{ l | ccc | ccc }
    \hline
    \multirow{2}{*}{Model} & \multicolumn{3}{c|}{NEEL Test} & \multicolumn{3}{c}{TACL-IE}   \\
                        & P & R & F1 & P & R & F1  \\ \hline
  Structured Perceptron & 74.1 & 59.5 & 66.0 & 59.4 & 54.7 & 56.9 \\
  Linear SSVM           & 72.3 & 65.5 & 68.8 & 61.0 & 63.0 & 62.0 \\ \hline
  Polynomial SSVM       & 68.0 & \tb{76.4} & 72.0 & 58.4 & 66.0 & 62.0 \\
  MART                  & 76.5 & 74.2 & 75.3 & \tb{61.5} & 67.0 & \tb{64.1} \\
  \sysname              & \tb{80.2} & 75.4 & \tb{77.7} & 60.1 & \tb{67.7} & 63.6 \\ \hline
  + entity-entity       & \underline{82.1} & 72.8 & 77.2 & \underline{67.1} & 67.3 & \underline{67.2}  \\
    \hline
\end{tabular}
\caption{IE-driven evaluation results on the full NEEL Test and TACL-IE datasets for different models. The best results with basic features are in \tb{bold}. The results are \underline{underlined} if adding entity-entity features gives the overall best results.}
\label{tab:full-results}
\end{table*}

The statistics of the full NEEL and TACL-IE datasets are shown in Table~\ref{tab:full-data}. Table~\ref{tab:full-results} presents the IE-driven evaluation results on the datasets. On the NEEL Test dataset, non-linear models significantly outperform linear models, tree-based models perform much better than alternative methods, and our tree-based structured prediction method \sysname~achieves the best results. The tree-based non-structured model MART yields best performance on the TACL-IE dataset with basic features, which slightly wins over \sysname~by 0.5\% F1. The entity-entity relationships improve the performance of \sysname~on the TACL-IE dataset by 3.6 points of F1. 

\end{appendices}

\end{document}